# A Spatio-Temporal based Frame Indexing Algorithm for QoS Improvement in Live Low-Motion Video Streaming

Adewale Emmanuel Adedokun, Muhammed Bashir Abdulrazak, Muyideen Momoh Omuya, Habeeb Bello-Salau, Bashir Olaniyi Sadiq
*Department of Computer Engineering,*
*Ahmadu Bello University, Zaria, Nigeria*

**ABSTRACT**
*Real-time video life streaming of events over a network continued to gain more popularity among the populace. However, there is need to ensure the judicious utilization of allocated bandwidth without compromising the Quality of Service (QoS) of the system. In this regard, this paper presents an approach based on spatio-temporal frame indexing that detects and eliminate redundancy within and across captured frame, prior transmission from the server to clients. The standard and local low motion videos were the two scenarios considered in evaluating the performance of the proposed algorithm. Results obtained showed that the proposed approach achieved an improvement of 5.13%, 15.8% and 5%, 15.6% improvement in terms of the buffer size and compression ratio. Though with a tradeoff of the frame-built time, where both the standard and local frame indexing outperforms the proposed scheme with 10.8% and 8.71% respectively.*

## INTRODUCTION

Videos comprises of multiple moving frame images recorded over a certain period of time with quick transition (Apostolopoulos *et al*., 2002; Vinod *et al*., 2014; Mahini *et al.,* 2017). The advent of the internet has revolutionized and ease the process of transmitting and broadcasting events in real time over the network (Gangurde & Nikam, 2017). In live streaming videos over the network, several applications have been developed among which include QuickTime, ActiveMovie, and RealPlayer (Kumar & Kumar, 2016). Though these applications were able to live stream videos. However, of importance is the need to ensure efficient utilization of the allocated network bandwidth that enable seamless streaming of the video with less computational time, low delay, during the encoding, transmission and streaming of the video.

Video streaming can simply be described as the process of viewing live event or recorded video media data directly over the network (Azhar *et al*., 2016). Several video image compression techniques have been proposed in literature (Kumar & Kumar, 2016; Apostolopoulos *et al*., 2002; Ma *et al.,* 2017) towards reducing the size of the moving image frame transmitted over the network for remote access by users via streaming in either real time or recorded access. Nevertheless, a major challenge associated with this technique is its computational complexity as well as the large dynamic image output, despite the application of the compression algorithm, which resulted into the inefficient utilization of the allocated bandwidth. Thus, this motivates the development of a spatio-temporal indexing algorithm proposed in this paper.

The major contribution of this paper is the development of a dynamic image compression algorithm based on spatio-temporal indexing, towards optimal bandwidth utilization of streaming network during real-time video streaming. The proposed algorithm comprises of two-fold approach namely, the server and the client module. The indexing of the image was carried out at the server end, where the first is set as reference and is sent to the client. The client end compared each subsequent pixel to the reference frame and detect pixels values that differs from the reference frame, compress these values along with the indexing information that is required for the reconstruction of the new frame at the client end. The reconstruction algorithm is implemented in the client end, which receives three important information. Firstly, the reference frame that serve as reference to subsequent frame that was built, secondly, is the difference buffer and lastly, the index buffer, which are all in compressed form. All the compressed information is being decompress with the frame reconstructed and corresponding video are being displayed to the remote user. These processes continue until the live-streaming end. Thereby, eliminating the redundancy that exists within and across frames. Consequently, reducing the latency, resolution degradation and packet loss that affect the QoS of video live streaming.

## RELATED LITERATURE

Major demerit with most documented approaches for video compression it's their complexity, which resulted in more processing and computational time. Thus, leading to high bandwidth requirements and computational time. In this regard, an approach based on optimal cache algorithm and priority selected cache algorithm for addressing the challenge associated with bandwidth utilization during video streaming as well as ensuring the quality of video playback was presented in (Chang *et al*., 2007). Experimental results obtained showed that the proposed approach was able to guarantee the quality of the video playback as well as minimized packet loss. However, the use of proxy makes the approach not applicable for bandwidth and time sensitive applications. Similarly, an efficient VoD streaming algorithm for broadband access network was presented in (Choi *et al*., 2010). The proposed approach utilized the video adaptive streaming, video greedy adaptive streaming and video greedy adaptive streaming with proactive buffering algorithm. These ensured the optimal utilization of the network bandwidth and the storage facilities during streaming with specific application to a passive optimal network. Results obtained showed that the average waiting time as well as request made by users were reduced. Thus, the network bandwidth consumption. However, the implementation of the proposed scheme was not considered for real-time video streaming.

Shen *et al*., (2013) implemented a scalable peer-peer live video streaming based on DHT-aided chunk driven overlay. It was observed that the tree-based peer to peer are vulnerable to churn, though the mesh-based system suffers high delay and overhead but resilient to churn. A video provider selection algorithm that aids the proper utilization of the system bandwidth, and a chunk sharing scheme that provides service for chunk index collection and discovery were used to guarantee high availability and utilization of the system bandwidth. While, system scalability was achieved by the DHT based infrastructure. Results obtained demonstrates an improved performance of the proposed scheme in terms of latency, availability, scalability and memory overhead compared to tree and mesh-based systems. Nevertheless, it requires high computational time and implementation cost.

A novel approach for improving the QoS in live video streaming was proposed by (Zhang *et al*., 2015). The proposed scheme was called guarantee mechanism of contingency resource (GMCR). A contingency server that provide contingency service for urgent chunks towards ensuring its timely arrival at the users' buffer was deployed. In implementing the GMCR, the chunk scheduling mechanism was adjusted. While partitioning the yet to delivered chunk into urgent and non-urgent based on their playback deadline. The peers request for the non-urgent chunks was based on P2P paradigm. The contingency server provides service for those chunks (non-urgent chunks) that have exceeded the playback deadline to become urgent chunks. Also, a queuing model was used in analyzing the quantitative relation between the amount of contingency server resources and the level of user's QoS in a peer to peer live streaming. However, the approach is not suitable for time and bandwidth sensitive application due to the use of contingency server.

An efficient application layer handover scheme that support seamless mobility for P2P live streaming utilizing multichannel communication was presented in (Parodkar and Bade, 2015). The signal strength (RSSI) of the access points (APs) was used in predicting the handover. A preset threshold value was used in comparing the computed difference in the measure RSSI signal between the current and the target AP and used by mobile peer to predict handover. Once the handover is predicted, neighbor peers transmit data to the mobile peer at a faster speed by switching their transmission mode. The data unit for data delivery and display was a video block. Each video was divided into small blocks, which are distributed to other peers through the mesh structure. Each peer display video after buffering and sequencing received blocks in memory. The peers exchange distributed video with each other on virtual overlay network. Experimental result show that P2P live streaming system considerably improve the playback continuity significantly as the number of participating peers increases. Nevertheless, the handover mechanism hinder the performance of the system at low bandwidth hindering the required QoS for time and bandwidth sensitive application.

Kumar and Kumar (2016) implemented an index base streaming algorithm for enhancing the quality of service in live video streaming for low-motion and ensure optimized bandwidth utilization. A reconstruction algorithm was situated at the client end, which receives three essential information namely, the first reference frame used in building the successive frame, the difference buffer and the index buffer in their compressed form. The reconstruction algorithm decompressed all the information and builds the frame and display the video to the user. The goal was to ensure that only pixels that differ from subsequent one is sent and reconstructed at the clients end instead of sending the entire frame. Despite the use of the indexing based streaming algorithm, the difference buffer was still observed to contained redundant pixels that can further be reduced by exploiting both the spatial and temporal redundancy simultaneously in order to achieve higher compression efficiency.

A Data Distribution Service (DSS) middleware for live streaming of full motion video was presented in (Al-Madani *et al*., 2017). The proposed DDS utilized its high-performance data delivery mechanism to improve the quality of video streaming over a network. The DDS unlike the conventional client/server model was developed based on publish/subscribe pattern. Network congestion over the network was mitigated by the deadline QoS policy set-up by the middleware layer. As an illustration, if the server subscriber waiting time for the net packet exceeds a certain predefined deadline, a notification message is sent to the subscriber to initiate a minimization of the codec rate to avoid congestion on subsequent streams. While, application load at the subscriber side were minimized using a time-based filtering policy. The proposed approach showed a great potential in distributing video over networks due to its inherent advantages of low bandwidth consumption, minimal packet loss and jitter. Nevertheless, it required high implementation cost.

*Corresponding author: Momoh, M. O.   ✉ momuyadeen@gmail.com ✉ Dept. of Computer Engineering, Ahmadu Bello University, Zaria. © 2019 Faculty of Technology Education, ATBU Bauchi. All rights reserved

Canel *et al.*, (2018) proposed an algorithm for identifying interested events in a video frames based on semantic content obtained using Deep Neural Network (DNN). The algorithm uses a hierarchy of filters to trade-off between end-to-end latency and aggressive decimation. The semantic diversity of the selected frame was maximized to handles bursty event in videos streaming.  The proposed approach has the merit of exact selection of desired number of relevant frames, creating a uniform output frame rate from a non-uniform streaming of interesting frames. This is in contrast to a more dynamic algorithm that adjust some selected frames based on the content of the video. Although, the algorithm succeeds in its requirement of achieving the exact desired reduction factor. However, the characteristic of the result depends on the size of the buffer. Using a very large buffer results in delay and memory overhead due to the polynomial complexity of the longest path algorithm.

**METHODOLOGY**

This section presents the proposed methodology for the development of the spatio-temporal frame indexing algorithm. This involve a two-fold implementation approach viz: the server and client's module. The server module comprises of the indexing algorithm, while the reconstruction algorithm was implemented in the client module. Details description of each stages is presented herewith

*Indexing Algorithm (Server Module)*

The indexing algorithm was implemented on the server machine, where the camera was connected. The first live capture frame from the camera was sent to the clients as a reference frame utilizing the developed indexing algorithm. Subsequent frames from the camera were scan towards identifying pixels value that differs (spatial and temporal differences) from the reference frame. These values are compressed alongside their indexing information needed for the reconstruction of each subsequent frame received at the client's end. We note that each new reconstructed frame at the client end serve as a reference frame for the next subsequent frame to be received. This process continued till the end of the live streaming. The overall indexing process is summarized in Fig. 1. While, the indexing value and their translation are summarized in Table 1. The first frame sent by the camera from the server to client is summarized in Fig 2. The compressed received frame 1 by the clients serves as the referenced frame at the client end for the second transmitted frame from the camera. Fig. 3 presents the received second transmitted frame from the camera. The shaded portion in the second captured frame represented the spatial and temporal changes in the frame when compared to the reference frame 1. This observable difference is extracted by the developed algorithm and stored in the difference buffer as shown in Fig. 4.

*Corresponding author: Momoh, M. O.  momuyadeen@gmail.com  Dept. of Computer Engineering, Ahmadu Bello University, Zaria.*

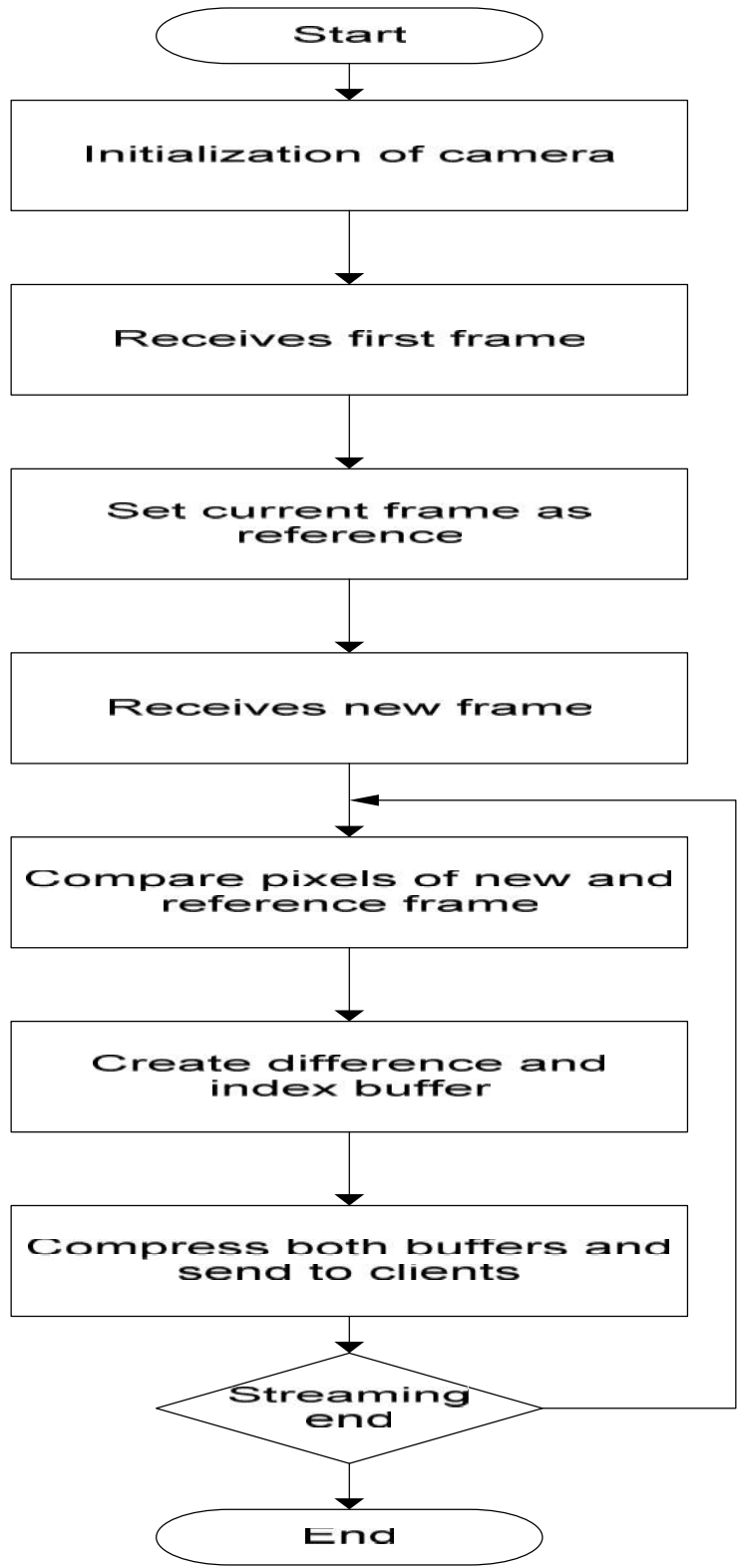

**Fig. 1:** Proposed indexing algorithm

*Corresponding author: Momoh, M. O. ✉ momuyadeen@gmail.com ✉ Dept. of Computer Engineering, Ahmadu Bello University, Zaria.* 

Fig 5. Presents the indexing buffer and its translational value has been summarized in Table 1.

**Table 1:** the indexing value and its translation

| Indexing value | Translation |
| --- | --- |
| -1 | Two successive equal frames |
| -2 | Copy pixel values from difference buffer |
| -3 | Copy pixel values from reference frame |
| -5 | Copy pixel value from difference buffer and replicate for x numbers of time |

**The Reconstruction Algorithm (Client Module)**

The reconstruction algorithm is situated at the client end which is used in reconstructing the transmitted message from the camera at the server module. Basically, three essential compressed information

| 156 | 159 | 158 | 154 | 151 | 152 | 156 | 154 | 152 | 156 |
| --- | --- | --- | --- | --- | --- | --- | --- | --- | --- |
| 156 | 162 | 158 | 159 | 161 | 149 | 149 | 141 | 153 | 154 |
| 156 | 157 | 158 | 157 | 155 | 154 | 159 | 157 | 155 | 154 |
| 152 | 152 | 152 | 153 | 154 | 151 | 156 | 151 | 153 | 154 |
| 152 | 151 | 150 | 153 | 157 | 151 | 159 | 158 | 159 | 154 |
| 155 | 154 | 152 | 150 | 149 | 149 | 149 | 149 | 151 | 154 |
| 156 | 159 | 158 | 159 | 157 | 154 | 146 | 148 | 147 | 150 |

Fig 2: First frame received by the client

| 156 | 159 | 158 | 154 | **155** | **155** | **155** | **155** | 152 | 156 |
| --- | --- | --- | --- | --- | --- | --- | --- | --- | --- |
| 156 | 162 | **157** | **157** | **157** | **157** | **158** | **159** | **158** | **156** |
| **155** | **158** | 155 | 157 | 155 | 154 | 159 | **153** | **152** | **156** |
| **156** | **156** | **156** | **156** | **158** | **159** | **154** | **152** | **150** | **150** |
| 150 | 151 | 150 | 153 | 157 | 151 | 159 | 158 | 159 | 154 |
| 155 | 154 | 152 | 150 | 149 | 149 | 149 | 149 | 151 | 154 |
| 156 | 159 | 158 | 159 | 157 | 154 | 146 | 148 | 147 | 150 |

Fig. 3: Second frame received by the client

| 155 | 157 | 158 | 159 |
| --- | --- | --- | --- |
| 158 | 156 | 155 | 158 |
| 153 | 152 | 156 | 158 |
| 159 | 154 | 152 | 150 |

Fig. 4: Difference buffer

| -3 4 | -5 4 | -3 4 | -5 4 | -2 6 | -3 5 | -2 2 | -5 5 | -2 4 | -5 3 | -3 29 |
| --- | --- | --- | --- | --- | --- | --- | --- | --- | --- | --- |

Fig. 5: The index buffer using the spatio-temporal frame indexing techniques

This is needed for the proper reconstruction of the frame during the live streaming of video are the reference frame, the difference and the index buffer. The reconstruction algorithm receives these information, decompress and reconstruct then display to the client and set as reference for the next frame.

The reconstruction process by the developed algorithm implemented at the client machine is summarized in Fig 6. Observe from Fig. 4 that the first index value which is -3 4, implies that the new frame in Fig 3 comprises of first four pixels which are the same when compared to the reference fame (Fig 2). Therefore, the first four pixels that were with that of the reference frame will be extracted from the reference frame and use for the reconstruction of the frame. Fig 7 presents the constructed frame after the first indexing value. The second index value -5 4, signifies that the next four pixels values differs from the reference but spatially related (see Fig 5). This means that the next pixel value can be copied from the difference buffer and duplicate four times. After the second indexing value, the construction of the frame is shown in Fig 8. The third index value -3 4 (see Fig 5) implies that the next four frames are same as the reference frame, this means that the next four pixels value can be copied from the reference frame. Fig 9 shows the construction of the frame after third indexing buffer. The fourth index value -5 4, signifies that the next four pixels values differs from the reference but spatially related (see Fig 5). This means that the next pixel value can be copied from the difference buffer and duplicate four times. The construction of the frame after the fourth indexing value is shown in Fig 10.

| 156 | 159 | 158 | 154 |  |  |  |  |  |
|---|---|---|---|---|---|---|---|---|
|  |  |  |  |  |  |  |  |  |
|  |  |  |  |  |  |  |  |  |
|  |  |  |  |  |  |  |  |  |
|  |  |  |  |  |  |  |  |  |
|  |  |  |  |  |  |  |  |  |
|  |  |  |  |  |  |  |  |  |

Fig 7. Construction of the frame after the first index value

| 156 | 159 | 158 | 154 | 155 | 155 | 155 | 155 |  |
|---|---|---|---|---|---|---|---|---|
|  |  |  |  |  |  |  |  |  |
|  |  |  |  |  |  |  |  |  |
|  |  |  |  |  |  |  |  |  |
|  |  |  |  |  |  |  |  |  |
|  |  |  |  |  |  |  |  |  |
|  |  |  |  |  |  |  |  |  |

Fig 8. Construction of the frame after the second index value

| 156 | 159 | 158 | 154 | 155 | 155 | 155 | 155 | 152 | 156 |
|---|---|---|---|---|---|---|---|---|---|
| 156 | 162 | 157 | 157 | 157 | 157 |  |  |  |  |
|  |  |  |  |  |  |  |  |  |  |
|  |  |  |  |  |  |  |  |  |  |
|  |  |  |  |  |  |  |  |  |  |
|  |  |  |  |  |  |  |  |  |  |
|  |  |  |  |  |  |  |  |  |  |

Fig 10. Construction of the frame after the fourth index value

| 156 | 159 | 158 | 154 | 155 | 155 | 155 | 155 | 152 | 156 |
|---|---|---|---|---|---|---|---|---|---|
| 156 | 162 | 157 | 157 | 157 | 157 | 158 | 159 | 158 | 156 |
| 155 | 158 |  |  |  |  |  |  |  |  |
|  |  |  |  |  |  |  |  |  |  |
|  |  |  |  |  |  |  |  |  |  |
|  |  |  |  |  |  |  |  |  |  |
|  |  |  |  |  |  |  |  |  |  |

Fig 11. Construction of the frame after the fifth index value

 

| 156 | 159 | 158 | 154 | 155 | 155 | 155 | 155 | 152 | 156 |
|---|---|---|---|---|---|---|---|---|---|
| 156 | 162 | 157 | 157 | 157 | 157 | 158 | 159 | 158 | 156 |
| 155 | 158 | 155 | 157 | 155 | 154 | 159 |  |  |  |
|  |  |  |  |  |  |  |  |  |  |
|  |  |  |  |  |  |  |  |  |  |
|  |  |  |  |  |  |  |  |  |  |
|  |  |  |  |  |  |  |  |  |  |

Fig 12. Construction of the frame after the sixth index value

| 156 | 159 | 158 | 154 | 155 | 155 | 155 | 155 | 152 | 156 |
|---|---|---|---|---|---|---|---|---|---|
| 156 | 162 | 157 | 157 | 157 | 157 | 158 | 159 | 158 | 156 |
| 155 | 158 | 155 | 157 | 155 | 154 | 159 | 153 | 152 |  |
|  |  |  |  |  |  |  |  |  |  |
|  |  |  |  |  |  |  |  |  |  |
|  |  |  |  |  |  |  |  |  |  |
|  |  |  |  |  |  |  |  |  |  |

Fig 13. Construction of the frame after the seventh index value

| 156 | 159 | 158 | 154 | 155 | 155 | 155 | 155 | 152 | 156 |
|---|---|---|---|---|---|---|---|---|---|
| 156 | 162 | 157 | 157 | 157 | 157 | 158 | 159 | 158 | 156 |
| 155 | 158 | 155 | 157 | 155 | 154 | 159 | 153 | 152 | 156 |
| 156 | 156 | 156 | 156 |  |  |  |  |  |  |
|  |  |  |  |  |  |  |  |  |  |
|  |  |  |  |  |  |  |  |  |  |
|  |  |  |  |  |  |  |  |  |  |

Fig 14. Construction of the frame after the eight index value

| 156 | 159 | 158 | 154 | 155 | 155 | 155 | 155 | 152 | 156 |
|---|---|---|---|---|---|---|---|---|---|
| 156 | 162 | 157 | 157 | 157 | 157 | 158 | 159 | 158 | 156 |
| 155 | 158 | 155 | 157 | 155 | 154 | 159 | 153 | 152 | 156 |
| 156 | 156 | 156 | 156 | 158 | 159 | 154 | 152 |  |  |
|  |  |  |  |  |  |  |  |  |  |
|  |  |  |  |  |  |  |  |  |  |
|  |  |  |  |  |  |  |  |  |  |

Fig 15. Construction of the frame after the ninth index value

| 156 | 159 | 158 | 154 | 155 | 155 | 155 | 155 | 152 | 156 |
|---|---|---|---|---|---|---|---|---|---|
| 156 | 162 | 157 | 157 | 157 | 157 | 158 | 159 | 158 | 156 |
| 155 | 158 | 155 | 157 | 155 | 154 | 159 | 153 | 152 | 156 |
| 156 | 156 | 156 | 156 | 158 | 159 | 154 | 152 | 150 | 150 |
| 150 | | | | | | | | | |
| | | | | | | | | | |
| | | | | | | | | | |

Fig 16. Construction of the frame after the tenth index value

| 156 | 159 | 158 | 154 | 155 | 155 | 155 | 155 | 152 | 156 |
|---|---|---|---|---|---|---|---|---|---|
| 156 | 162 | 157 | 157 | 157 | 157 | 158 | 159 | 158 | 156 |
| 155 | 158 | 155 | 157 | 155 | 154 | 159 | 153 | 152 | 156 |
| 156 | 156 | 156 | 156 | 158 | 159 | 154 | 152 | 150 | 150 |
| 150 | 151 | 150 | 153 | 157 | 151 | 159 | 158 | 159 | 154 |
| 155 | 154 | 152 | 150 | 149 | 149 | 149 | 149 | 159 | 154 |
| 156 | 159 | 158 | 159 | 157 | 154 | 146 | 148 | 147 | 150 |

Fig 17. Construction of the frame after the eleventh index value

From Fig. 5, the -2 6 means that the next six pixels value after the construction of the fourth index value differs from the reference frame and as such can be copied from the difference buffer. Fig. 11 shows the construction after the fifth index values.

The sixth index value -3 5 (see Fig 5) implies that the next five pixels value are same as the reference frame, this means that the next five pixels can be copied from the reference frame. Fig 12 presents the construction after the fifth index value. Observe from Fig 5 that the seventh indexing value -2 2, implies that the next two pixels differs from the reference frame and as such can be copied from the difference buffer. The construction of the frame after the seventh indexing value is presented in Fig 13. The eight index value -5 5, signifies that the next five pixels values differs from the reference but spatially related (see Fig 5). This means that the next pixel value can be copied from the difference buffer and duplicate five times. The construction of the frame after the eight indexing value is shown in Fig 14. The ninth index value -2 4 (see Fig 5) means that the next four pixels differs from reference frame and as such can be copied from the reference frame. Fig 15 presents the reconstruction of the frame after the ninth indexing value. Observe from Fig 5 that the tenth indexing value -5 3, implies that the next three pixels differs from the reference frame but spatially related and as such can be copied from the difference buffer and duplicate three times. The construction of the frame after the tenth indexing value is shown in Fig 16. The final value in the index buffer -3 29 (see Fig 5) indicates that next last twenty nine pixels value are same with the reference and can be copied from the reference frame. Fig 17 present the final frame after the last indexing value.

*Corresponding author: Momoh, M. O.  ✉ momuyadeen@gmail.com ✉ Dept. of Computer Engineering, Ahmadu Bello University, Zaria.

**SIMULATION RESULTS AND DISCUSSION**

This section presents the simulation results obtained when the proposed algorithm was applied to a local low motion video. The performance of the proposed algorithm was compared to a similar algorithm proposed in (Kumar & Kumar, 2016) in terms of three different metrics namely, difference buffer size, compression ratio and time to build up exploiting the spatial and temporal correlation. We note that pixels values that differ from the previous frame along with the indexing information that are necessary for the reconstruction are sent from the server camera to the client. Each reconstructed frame information as a reference to the next subsequent frame that will be received from the client. This process continued till the end of the live streaming.

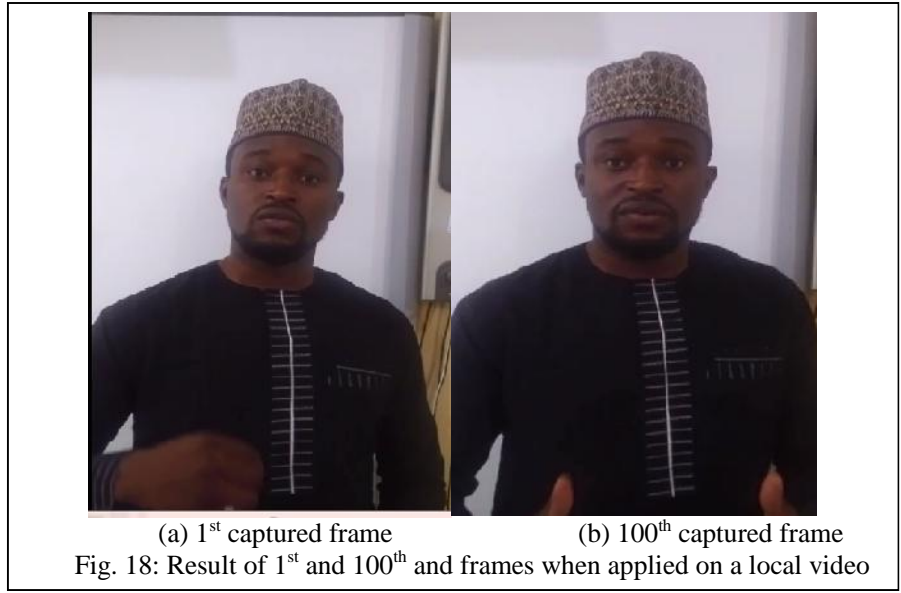

(a) 1$^{st}$ captured frame      (b) 100$^{th}$ captured frame
Fig. 18: Result of 1$^{st}$ and 100$^{th}$ and frames when applied on a local video

frame. The 1$^{st}$ and 100$^{th}$ frame captured during the implementation of the proposed algorithm on a local low motion video are presented in Fig. 18.

Observe, that during the streaming process each captured frame from server are sent to the clients which compared each received frame with the immediate passed received (reference) frame until the whole frame are reconstructed at the client end by Thus, minimizing redundancy in each successive frame. consequently, ensuring the optimal bandwidth utilization during live streaming and improve QoS as end to end delay, packet loss, and resolution degradation are minimized. Some randomly selected frames extracted from the local low-motion video from Fig 18 with an initial frame size of 373 X 597 X 3, in terms of buffer size is presented in Table 2.

**Table 2:** Comparison of the buffer size of some randomly selected frames in the sample video

| Number of frames | Standard indexing (%) | Spatio-temporal indexing (%) |
|---|---|---|
| 4 | 31.8470 | 27.7311 |
| 11 | 36.6630 | 35.8371 |
| 14 | 24.2988 | 20.7534 |
| 20 | 28.1396 | 25.9823 |
| 33 | 39.9416 | 32.6465 |
| 40 | 40.6100 | 38.8384 |
| 54 | 27.1202 | 24.0001 |
| 60 | 27.6670 | 26.6515 |
| 100 | 41.2634 | 31.7217 |

*Corresponding author: Momoh, M. O. ✉ momuyadeen@gmail.com ✉ Dept. of Computer Engineering, Ahmadu Bello University, Zaria. 

| | | |
|---|---|---|
| 123 | 23.2913 | 15.2237 |

Observe that streaming using spatio-temporal frame indexing techniques requires a lesser size of difference buffer compared to that of standard frame indexing techniques on a local low-motion video (See Table 2). The percentage of pixels that was used for the reconstruction of the video using the spatio-temporal frame indexing algorithm was 27.61% while 32.80% was used for reconstruction using the standard frame indexing algorithm. Similarly, a slight increase of about 2.09% was observed in the frame build up time using the developed spatio-temporal frame indexing algorithm compared to the standard frame indexing (See Table 3). This can be associated to increase in the indexing information required for the reconstruction of frames.

**Table 3:** Comparison using frame built time of the standard and the spatio-temporal frame indexing

| Number of frames | Standard indexing (sec) | Spatio-temporal indexing (sec) |
|---|---|---|
| 4 | 0.0235 | 0.0237 |
| 11 | 0.0245 | 0.0259 |
| 14 | 0.0290 | 0.0318 |
| 20 | 0.0125 | 0.0131 |
| 33 | 0.0160 | 0.0166 |
| 40 | 0.0177 | 0.0215 |
| 54 | 0.0204 | 0.0214 |
| 60 | 0.0155 | 0.0185 |
| 100 | 0.0178 | 0.0185 |
| 123 | 0.0181 | 0.0190 |

We note that the proposed scheme achieves an improved compression efficiency compared to the existing standard scheme when implemented on a local low-motion video with an average percentage improvement of about 15.6% (See Table 4)

Table 4 Compression ratio of some randomly selected frame in the sample video

| Number of frames | Standard indexing (%) | Spatio-temporal indexing (%) |
|---|---|---|
| 4 | 0.32 | 0.28 |
| 11 | 0.37 | 0.36 |
| 14 | 0.24 | 0.21 |
| 20 | 0.28 | 0.26 |
| 33 | 0.40 | 0.30 |
| 40 | 0.41 | 0.39 |
| 54 | 0.27 | 0.24 |
| 60 | 0.28 | 0.27 |
| 100 | 0.41 | 0.32 |
| 123 | 0.23 | 0.15 |

**CONCLUSION**

A spatio-temporal frame indexing algorithm for improving the QoS in live low-motion video streaming is presented in this paper. The proposed spatio-temporal frame indexing algorithm was developed by exploiting both the spatial and temporal correlations in each frame and the next successive frames as opposed the conventional streaming approach where the entire frame are transmitted by exploiting only the temporal correlations. It was noted that, only the dissimilar pixels are transmitted along with their buffers information using the proposed scheme. The reconstruction of the transmitted frame from the server occur at the client end using the difference pixels and the buffers information. This aid in judicious utilization of the network bandwidth during the live video streaming as well as minimizing traffic in the network during transmission and streaming. The proposed algorithm recorded an improved performance in terms of the buffer size and compression ratio when compared to the standard frame indexing. Though, at the expense of slight increment in the required time to build. Future research work will explore approaches of improving the required time needed to build or reconstruct frames. In addition, the possibility of extending the proposed approach for streaming to high motion video will be consider.

*Corresponding author: Momoh, M. O. ✉ momuyadeen@gmail.com ✉ Dept. of Computer Engineering, Ahmadu Bello University, Zaria.

*Corresponding author: Momoh, M. O.   ✉ momuyadeen@gmail.com  ✉ Dept. of Computer Engineering, Ahmadu Bello University, Zaria.